\documentclass[letterpaper, 10pt, conference]{ieeeconf}

\renewcommand{\baselinestretch}{.964}
\IEEEoverridecommandlockouts

\title{\LARGE \bf
Multi-Arm Bin-Picking in Real-Time:\\ A Combined Task and Motion Planning Approach
}

\author{
Ilyes Toumi$^{{1},{3}}$,  
Andreas Orthey$^{2}$, 
Alexander von Rohr$^{3}$ and
Ngo Anh Vien$^{1}$
\thanks{\small $^{1}$ Bosch Center for Artificial Intelligence}
\thanks{\small $^{2}$ Technical University of Berlin}%
\thanks{\small $^{3}$ Institute for Data Science in Mechanical Engineering, RWTH Aachen University, Germany}%
}

\makeatletter
\let\NAT@parse\undefined
\makeatother
\usepackage[numbers]{natbib}
\usepackage{graphicx}

\graphicspath{{figures/}}
\usepackage{caption}
\usepackage{subcaption}
\usepackage{comment}

\usepackage{xcolor}
\usepackage{amsmath}
\usepackage{amsfonts}
\usepackage{balance}
\usepackage{mathtools}
\usepackage{booktabs}
\usepackage{multirow}
\usepackage{natbib}

\begin{document}
\maketitle

\thispagestyle{empty}
\pagestyle{empty}

\begin{abstract}
Automated bin-picking is a prerequisite for fully automated manufacturing and warehouses. To successfully pick an item from an unstructured bin the robot needs to first detect possible grasps for the objects, decide on the object to remove and consequently plan and execute a feasible trajectory to retrieve the chosen object. Over the last years significant progress has been made towards solving these problems. 
However, when multiple robot arms are cooperating the decision and planning problems become exponentially harder.
We propose an integrated multi-arm bin-picking pipeline (IMAPIP), and demonstrate that it is able to reliably pick objects from a bin in real-time using multiple robot arms. IMAPIP solves the multi-arm bin-picking task first at high-level using a geometry-aware policy integrated in a combined task and motion planning framework. We then plan motions consistent with this policy using the BIT* algorithm on the motion planning level. We show that this integrated solution enables robot arm cooperation. In our experiments, we show the proposed geometry-aware policy outperforms a baseline by increasing bin-picking time by 28\% using two robot arms. The policy is robust to changes in the position of the bin and number of objects. We also show that IMAPIP to successfully scale up to four robot arms working in close proximity.

\end{abstract}

\section{Introduction}

\label{sec:INTRODUCTION}

\begin{figure}
 \centering
 \includegraphics[width=0.7\linewidth]{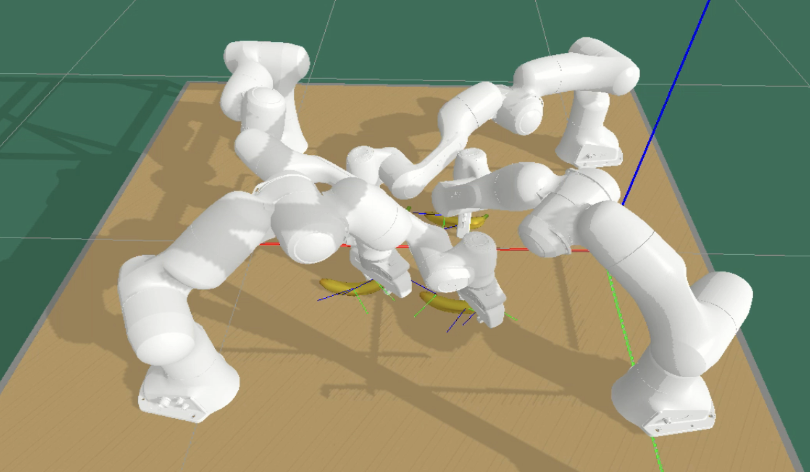}
 \caption{Example of four robots grasping objects}
 \label{fig:setup}
\end{figure}

In manufacturing and warehouse storage facilities, it is common for objects and workpieces  to be stored and transported in unstructured bins to save the cost of special containers or custom stacking methods. This makes subsequent automated removal of the workpieces a more complex task and often results in operations carried out manually, which is expensive and physically demanding work for humans. 
This problem can be automated by using robot arms. However, the speed of work is generally limited by the number of robots used which is in the vast majority of the cases limited to one robot which makes the idea to use multiple arms interesting. Due to the real-time requirements most prior work has concentrated on bin-picking with one robot arm. This rare work \cite{schwarz2018fast}, discussed in \ref{sec:RELATEDWORK}, had attempted the dual-arm bin-picking application. However, some simplification and workaround were made. The main bottleneck of a multi-arm robotic system for bin picking applications lies in the computational cost for real-time motion planning and the lack of task coordination policies that assign and schedule the tasks to optimize bin picking rate.
As a summary, our main contributions are three-fold as follows:
\begin{itemize}
    \item The practical grasping pipeline that can robustly and simultaneously operate multiple robot arm for bin-picking applications in real-time. 
    \item A general task and motion planning (TAMP) formulation for multi-robot bin-picking where task planning handles robot-grasp decisions while motion planning plans motion for a system of multi robot at a geometric level. We propose an approximation solution to achieve the requirements of real-time computation. 
    \item The proposed approximation enables different ideas for task planning. Our proposed task policy, called the \emph{geometry-aware policy}, can robustly and accurately coordinate the tasks between two and four robots with the possibility to adapt to different scenarios of picking as illustrated in Fig. \ref{fig:setup}.

\end{itemize}



\section{Related Work}
\label{sec:RELATEDWORK}
Robot grasping has been a long-standing research problem for robotic manipulation. Therefore, it has a large body of literature.
\subsection{Robotic Bin-Picking}
\label{subsec:Binpicking}


There are a large amount of grasp learning papers on bin-picking \cite{abs-2207-02556}. However all these reviewed approaches only consider single-arm picking. For instance, \cite{kumra2020antipodal} present a generative residual convolutional neural network (GR-ConvNet) architecture that predicts pixel-wise antipodal grasp poses using RGB-D images input. 
Dex-Net \cite{mahler2019learning} addressed the limitations of single-arm antipodal grasping with an ambidextrous grasping where each robotic arm is equipped with either a suction or parallel jaw gripper. There has been work introducing a dual-arm robotic grasping system for the Amazon Robotic Challenge (ARC) by \cite{schwarz2018fast}. This system can perform online dual-arm coordination planning and execution control for picking tasks. However their approach uses a simple workspace separation strategy to allocate grasps to each robot. Therefore it would significantly reduce the overall task performance. Our proposed approach will address this issue in a principled way that helps both improving the task performance and scaling to more robots. 



\subsection{Multi-Arm Motion Planning}
\label{Motionplanning}
Multitude approaches were proposed to solve multi robotic planning. 
One approach is to consider a multi-arm robot system as one composite robot whose resulting robot DOFs is the sum of all robots DOFs, then apply standard sampling-based motion planning algorithms. The first application of this approach is from \cite{sanchez2002using} applied to a fleet of robots. This approach main's limitation is the computation cost that increase exponentially for each additional DOFs. A second approach use prioritization frameworks where each robot path is planned sequentially and using this path as a constrain for the next robot such as \cite{erdmann1987multiple,wu2020multi,bennewitz2001optimizing,grothe2022st}.


Despite making breakthroughs in the multi-arm motion planning problem, all these work are not specifically targeted for bin picking application. It is unclear which motion planner would work the best for the multiple arm robots in bin picking application. Therefore, we conduct our benchmark to answer this question.

\subsection{Task and Motion Planning}
\label{Taskplanning}
Combined task and motion planning (TAMP) designates the problem of sequential robot manipulation where the objective is primarily given in terms of cost function over the final geometric state \cite{toussaint2015logic,ToussaintMMWVL16}. This means that the robots and objects have to have a specific final reach position. 

The general approach of solving task and motion planning problems, according to \cite{garrett2020integrated}, can be either by combining the discrete logic search with a constraint satisfaction methods \cite{lagriffoul2014efficiently,lozano2014constraint,lagriffoul2012constraint} or with a sampling-based motion planning algorithm \cite{dantam2018incremental,de2013towards,kaelbling2010hierarchical}. The number of feasible symbolic sequences increases exponentially with the number of objects' sequence length. This leads to a large number of geometric problems that need to be solved. Therefore, heuristics are used to solve this problem, such as efficiently pruning the search tree. Another approach for TAMP is to combine the logic search with trajectory optimization \cite{toussaint2015logic}. Toussaint \cite{toussaint2015logic} proposes an optimization-based approach to combined task and motion planning called logic geometric programming (LGP) where the high-level idea is a nonlinear trajectory optimization problem over the continuous path of all robot joints and objects in the scene. The constraints and costs of this trajectory optimization problem are parameterized by a discrete variable representing the state of a symbolic domain.

In general, an optimization-based approach has solutions with trajectories that are optimized with global consistency 
\cite{hartmann2021long}. Heuristics can be introduced to speed up solution findings; whether hand crafted or learned, they still need to search over the discrete variables and probably solve many motion planning problems. One example of an heuristics, presented by \cite{kaelbling2010hierarchical,rodriguez2019iteratively}, consist of pruning the search tree. Another example is to integrate learning into TAMP to guide the discrete search \cite{chitnis2016guided,ChitnisSKKL20,garrett2016learning,wang2018active}. These approaches are mainly used for a single task with no possibility to generalize to different scenarios. 

Our approach directly extends those works to the problem of multi-arm robotic manipulation in bin-picking application. This is not straightforward, since we have to deal with worlds in which we need to operate in real-time, robustly generalize to any number of objects in the scene, and, most importantly, generalize to multiple robots.

\section{Multi-Arm Bin-Picking Problem}

In this section, we describe our proposed problem formulation. We propose to leverage the general formulation of TAMP under partial observability for multi-arm bin-picking.

\subsection{Problem Statement}
\label{sec:Mathematical_Statement}
We consider a multi-arm robotic bin-picking problem where there are a system of $N$ robots with configuration spaces ${\cal X}_1$, ..., ${\cal X}_N$ picking on a shared bin of a finite number of objects. The system is integrated with one or more overhead depth camera. At each open-loop grasp round $k$, the system captures a point cloud image $I$ of the scene as observation $y_k$ from the camera. The system then computes a unified policy $\pi(y_k) = \{\pi_r,x_r\}_{r=1}^N$ \footnote{We use subscripts $k$ and $r$ interchangeably on the same variable for different purposes, where we denote either grasp step index $k$ or robot index $r$, respectively.} to plan both task assignment $\pi_r$ and motion plan $x_r$ for each robot $r$. The sub-policy $a_k(r) = \pi_r(I)$ defines a grasp action for a gripper of robot $r$, e.g. a 3D rigid position and orientation of the gripper $T_g = (R_g, t_g) \in \text{SE}(3)$. The execution of action $a_k(r)$ is defined by following 
a grasp motion path computed as a mapping $x_r: [t_k,t_{k+1}]  \mapsto {\cal X}_r$, where ${\cal X}_r$ is a configuration space of robot $r$, and $[t_k,t_{k+1}]$ defines a time segment between grasp round $k$ and $(k+1)$. After $a_k$ is executed, each robot receives a reward $R_{r,k}=1$ if policy $\{\pi_r,x_r\}$ succeeds in lifting one object and dropping it to a targeted place; otherwise $R_{r,k}=0$. Our goal is to optimize a policy $\pi$ that maximizes an average total reward of all robots over $K$ grasp rounds.

\subsection{Problem Formulation}
\label{sec:Mathematical_formulation}
Bin-picking with only one robot arm is a simple manipulation setting, thus it requires fairly simple motion planning, e.g. a simple linear motion planner \cite{mahler2019learning}. 
In \emph{multi-arm bin-picking} settings, there are two sources of challenges that render all existing approaches for single-robot infeasible,
\begin{itemize}
    \item Complex task and motion planning: Robots are required to know which robot should grasp which object and plan the motion path accordingly while taking into account both collision avoidance with other robots and clearing the bin with the least number of pick attempts.
    \item Uncertainty: As defined in bin-picking literature \cite{mahler2019learning}, there are two type of uncertainties. First, the uncertainty of current-state about what the object states are, given point clouds as observation, e.g. center of mass, friction coefficients, and geometry poses. Second, the uncertainty of future-state about what the outcomes of all robots' actions possibly are.
\end{itemize}

We propose to address the above challenges by adopting the general integrated task and motion planning in belief space \cite{kaelbling2013integrated}. In our formulation, we propose grasp selection to be task planning in which a plan is defined as a mapping from an input image of the picking scene to an assignment of one grasp to each robot. Given point clouds as observation, object states are not fully observable, therefore an optimal task plan should be optimized over a belief space. Formally, we define this belief planning problem as a partially observable Markov decision process (POMDP) with tuple ($\mathcal{S}, \mathcal{A}, \mathcal{T}, \mathcal{O}, \phi, \mathcal{Y}$), where
\begin{itemize}
    \item \textbf{State} $s\in \cal S$ is a set of joint ground-truth information of the environment, i.e. all object's geometry, pose, center of mass, friction coefficients, all robots' states and all sensors' states.
    \item \textbf{Action} $a \in \mathcal{A}$ is a set of joint picking actions of all robots $a_k=\{a_r\}_{r=1}^N$, e.g. a picking pose to each robot.
    \item \textbf{Observation} $y \in \mathcal{Y}$, is a set of RGB-D image data. 
    \item \textbf{Transition} $\cal T$ defines a state transition given action $a$, i.e. ${\cal T}:{\cal S} \times {\cal S} \times {\cal A}\mapsto [0,1]$ where ${\cal T}(s',s,a) =p(s'|s,a)$. Variable $s'$ denotes the following state of $s$. 
    \item \textbf{Observation function} $\mathcal{O}$ defines an observation distribution given ground-truth state $s$, i.e. $\mathcal{O}(y,s)=p(y|s)$.
    \item \textbf{Task cost function} $\phi$ defines a cost given state $s$ and action $a$, i.e. grasp success or failure.
    \end{itemize}


Assuming there are $N$ robot arms $\{R_r\}_{r=1}^N$ where $r$ is the index of the robot. 
At a full state $s_k$, each robot decides to perform an action $a_r$, after which it transits to state $s_{k+1}$ while making the next observation $o_{k+1}$, e.g. the bin has fewer objects resulting in a next observation $o_{k+1}$. The target of the high-level decision process is to find an optimal policy $\pi$ which is a mapping from state space $\cal S$ to action space $\cal A$ that minimizes the total task cost, i.e. minimizing the total of grasp failures. 

The above decision making process is high-level, that provides only task decisions i.e., robot $r$ picks a particular object with one predicted grasp pose, etc.. However it does not define how motion at geometric-level can be realized. In trajectory optimization, we have to optimize the joint motion of the robots given the task constraints given by a joint action $a_k$. At geometric-level, each action $a_k$ within only interval $[t_k, t_{k+1}]$ corresponds a joint motion $x=\{x_r\}_{r=1}^N \in {\cal X}$ of all robots where ${\cal X} = \{{\cal X}_r\}_{r=1}^N$ is a joint configuration space, and $x_r \in {\mathcal X}_r$. This would require to solve a non-linear mathematical optimization problem as follows,
\begin{equation}
\begin{aligned}
&\min_{x} c (a_k, x) \\
&\text{s.t.} \,g_{a_k}(x(t), \dot{x}(t),\ddot{x}(t)) \le 0, 
\, h_{a_k}(x(t), \dot{x}(t),\ddot{x}(t))=0,
\end{aligned}  
\label{eq:motion}
\end{equation}
where the motion objective is $$c (a_k, x)=\int_{t_k}^{t_{k+1}} f_{a_k}(x(t), \dot{x}(t),\ddot{x}(t)) dt,  $$ and the functions $f_a(x(t), \dot{x}(t),\ddot{x}(t))$, $g_a(x(t), \dot{x}(t),\ddot{x}(t))$, and $h_a(x(t), \dot{x}(t),\ddot{x}(t))$ denotes the cost and constraints of the motion within the phase of joint action $a$.

We now combine the above two decision processes, i.e. task and motion level, into a single optimization. The idea is we fix a skeleton of joint actions, i.e. $a_{1:k}$ up to $k$ grasping rounds, and then optimize motion conditioned on the skeleton while taking into account beliefs over uncertainties. This POMDP-TAMP problem optimizes jointly a task policy $\pi:{\cal S} \mapsto {\cal A}$ and a joint motion plan $x:[0,1] \mapsto{\cal X}$ for a set of robot arms as follows
\begin{equation}
\begin{aligned}
&\pi, x = \arg\min_{\pi,x}  \sum_{k=0}^T \int_{t_k}^{t_{k+1}} Pr(x|\{y_{0:k}\})  c(a_k, x) dx\\ 
&\text{s.t.}\,  g_a(x(t), \dot{x}(t),\ddot{x}(t)) \le 0, \, h_a(x(t), \dot{x}(t),\ddot{x}(t))=0 ,
\end{aligned}
\label{eq:pomdp_tamp}
\end{equation}
where we denote $Pr(x|\{y_{0:k}\})$ as a belief distribution of all robots' state given a history of observations $\{y_{0:k}\}$ up to time $k$, and $T$ denotes the horizon length. A realization of $x$ at time $k$ will define the explicit form of the trajectory optimization for motion within interval $[t_k, t_{k+1}]$.

\subsection{Practical Aspects}
\label{relaxations}
Solving the above POMDP-TAMP problem in Eq. \ref{eq:pomdp_tamp} in a global manner can be intractable. Therefore we present assumptions and possible relaxations to reduce the computational complexity in addition to practical reasons. 

\begin{itemize}
    \item We assume an application of a pretrained grasp learning network which provides a set of grasp proposals $\cal A$, e.g. grasp poses, given an input image $y_k \in {\cal Y}$. In our work, we use the graph recognition convolutional neural network (GR-ConvNet) from \citeauthor{kumra2020antipodal} \cite{kumra2020antipodal}. The network receives a RGB-D image as input, and predict pixel-wise grasp quality and grasp configurations (gripper's angle and width). As a result, GR-ConvNet helps construct a finite set $\cal A$ of feasible grasps with high probability.
    \item In practice, bin-picking is a stochastic task which results in a long horizon POMDP-TAMP problem under partial observability where solving it becomes exponentially more complex. Since the main goal in bin-picking applications is to have a real-time computation, we propose to compute only a suboptimal myopic policy $\pi(y)$ with a horizon of $T=1$ \cite{krishnamurthy2009partially}. This myopic policy is often used and shown to be sufficiently effective in bin-picking use-cases \cite{mahler2019learning}.   
\end{itemize}

    
    
    
\subsection{Two-Level Approximation for POMDP-TAMP}
\label{sec:approximation}
Given the relaxation assumptions in Sec. \ref{relaxations}, our POMDP-TAMP is still computationally expensive to solve and does not meet real-time computation requirements. In particular, assuming that for each object in the scene of $n$ objects, we limit GR-ConvNet to propose $m$ grasp poses. Then the action space is a combinatoric number of the number of robots used in the system: $|{\cal A}|=(m \times n)^N$. Therefore, to select the best action even with a myopic policy would have to solve this TAMP optimization,
\begin{align}
\min_{a_0 \in {\cal A}, x \in {\cal X}} c(a_0,x),
\label{approx}
\end{align}
 $|{\cal A}|$ times, where $c(a_0,x)$ is defined in Eq.~\ref{eq:motion}. As a result, for real-time applications it does not scale to $N>1$.

Inspired by multi-level approximation approaches for TAMP \cite{Toussaint15,ToussaintL17}, we propose two levels of approximation for our simplified POMDP-TAMP as follow,
\begin{itemize}
\item The \emph{first level problem} ${\cal P}_1$ is coarse where action $a_0$ is selected using a sampling-based task planning approach. This task planning problem is defined as selecting the best action $a_0 \in {\cal A}$ that minimizes a task cost $\phi(\cdot)$. Our sampling-based approach samples a small set of $n \ll |{\cal A}|$ actions and choose an action with the best cost $\phi(\cdot)$ while not considering motion $x$. Specifically, we will discuss about different choices of the task cost $\phi(\cdot)$ in next section. Each choice results in a different policy of robot-grasp pose assignment, called {\bf geometry-aware policy}.
\item The \emph{second level problem} ${\cal P}_2$ is finest, which is the full motion planing problem Eq.~\ref{approx} with fine motion steps. This motion planning problem conditional to a given action $a_0$ found by the first level problem, which is an assignment of each robot to one grasp pose. A path plan $x$ is found by connecting grasp poses where we compute a joint grasp motion path for a multi-arm system. If an infeasible plan is returned, a next best action found by the coarser problem in the first level will continue to be checked.

\end{itemize}

\section{Integrated Multi-Arm Bin-Picking with Geometry-aware Policy}
\label{sec:IMPIP}

We present our approach, the Integrated Multi-Arm Bin-Picking pipeline using geometry-aware policy (IMAPIP). IMAPIP works by combining two main components which are an integrated bin picking system, we call grasping pipeline and our geometry-aware policy. 
\subsection{Integrated Multi-Arm Bin-Picking}
\label{sec:Graspingpipeline}
Our integrated bin-picking system is a robotic system composed of several modules that can grasp objects from inside a bin. This grasping pipeline autonomously senses its environment using an over-head RGB-D camera, detects and estimates pixel-wise grasp proposals, then selects the best grasp-robot assignment and optimizes motion for one robot, or multiple robots in our case. In particular, the camera placed on the top of the heap bin captures the scene as a RGB-D image input. Then it feeds it to the grasp recognition convolutional neural network (GR-ConvNet) from Kumra et. al. \cite{kumra2020antipodal}. GR-ConvNet predicts a pixel-wise grasp map in which each point in the grasp map estimates the position in the pixel coordinate, the roll around the \(z\) axis, the grasp width, and the grasp quality. From the pixel coordinate of an object, we estimate the object's position in the global coordinate using a transformation matrix from the pixel coordinates of the camera to the robot frame. From this transformed grasp pose, the joint position of the robot for the object to grasp is computed using inverse kinematic.

From the grasp map, we can build a list of $|\cal A|$ grasps with highest grasp quality. We denote this set as the action space $\cal A$ of the task planning problem ${\cal P}_1$, we solve the POMDP-TAMP problem in Eq.~\ref{approx} using the two-level approximation method as described in \ref{sec:approximation}. The found policy $a_0$ will assign the task for each robot, namely which robot should pick which object with which pose and in what order, then solve the problem ${\cal P}_2$ to plan the motion for all of them. 
At this step, the path of the robots are controlled using a linear proportional-derivative (PD) controller. The controller commands the robot to move to the grasp position, pick then place the object of interest in the goal position. For every successful pick and place, the cycle will repeat until the bin is cleared.

\subsection{Task Planning as Grasp Synthesis Policies}
In this part, we describe how the problem ${\cal P}_1$ as described in \ref{sec:approximation} is solved. IMAPIP uses different task costs $\phi(\cdot)$ for this task planning problem. Each task cost results in a different policy to dictate which robot should pick which object with which pose. We call these policies: a sequential policy (a simple baseline, a split space policy (a baseline), and geometry-aware policy. 

\subsubsection{Sequential policy}
This simple baseline will operate the robots sequentially, i.e., only one robot at a time while the other robot stays still in its home position and picks all reachable objects in an alternating manner. The choice of robot is random. The sequential policy can be considered to simplify the multi-arm system into a single-arm system and is commonly used in different academic work and industrial setups, for example \cite{mahler2019learning}. 

\subsubsection{Split space policy} This policy divides the workspace into the Voronoi regions, for example two separate workspace like in \cite{schwarz2018fast}. Depending on which region the object is located, it will be assigned to the corresponding closest robot. If two object lies too close to each other and can not be grasped simultaneously, the  robots will pick sequentially. 


\subsubsection{Geometry-aware policy} This type of policy is based on formulation which computes the best set of objects to grasp following three different criteria that are aware of: kinematics feasibility, robot-object geometric distance, and grasp quality. We first pre-process the set of grasp proposals $\cal A$ by filtering out unreachable grasps corresponding to each robot. This results in a bounded set of $n_1 \times n_2 \times \ldots \times n_N$ combinatoric set of possible object-grasp assignment, where $n_r$ represents the set of reachable grasps for robot $r$. We propose the following task cost $\phi(\cdot)$:

\begin{itemize}
\item Kinematic feasibility: This policy uses a task cost $\phi(a)=1$ if all actions $a_r$ are collision-free, otherwise 0; where $a=\{a_r\}_{r=1}^N \in {\cal A}$. 
\item Grasp quality aware: This policy uses as cost the sum of all grasp qualities (as output of GR-ConvNet) as encoded in the robot-object assignment action $a$ as task cost $\phi(\cdot)$.
\item Distance aware policy: This policy uses the sum of pair-wise Euclidean distance between grasps as encoded in the robot-object assignment action $a$ as task cost $\phi(\cdot)$.
\end{itemize}

The obvious limitation of the kinematic feasibility policy is that it does not tell us much about the best pair of object to grasp in terms of grasp quality or safety to collision. In contrary, the grasp quality aware and distance aware policies help perform high quality grasp or improve motion planning computational time by taking the distance between final grasp poses into account. In our implementation, for kinematic feasibility policy we did not select a fixed number $n$, instead we randomly sample an action $a\in{\cal A}$ until it satisfies $\phi(\cdot)=0$, because the cost of the collision check is expensive. For the other two policies we use them only in the setting of two-arm picking. Therefore we use a full filtered action space $n=n_1\times n_2$ because its computation cost is cheap. If no feasible actions are found, the robots will operate in a sequential manner. From this point we will refer to it as the sequential policy.




\subsection{Motion Planning}
To solve the second level problem in \ref{sec:approximation}, IMAPIP uses different off-the-shelf motion planning methods. Due to the planners' unique characteristics, the motion planners used in IMAPIP can be divided into two main categories: Parallel planning and prioritization framework. Parallel planning consists of considering all involved robots as one composite robot, for instance the dual-arm robot system of 7 DOF Franka-Emika panda will be considered as 14 DOF. Parallel planning will be used, for RRT  \cite{lavalle2001randomized}, RRT* \cite{karaman2011sampling}, RRT-Connect \cite{kuffner2000rrt}, {ABITstar} \cite{strub2020advanced}, BIT* \cite{GammellSB14a}. The prioritization framework will generate a path for the first robot and then use it as a constrain for the next robot. This approach will be used to test the ST-RRT \cite{grothe2022st}. The selection of motion planners comes from two key criteria: optimality and bi-directionality. 

We conducted a benchmarking experiment of these motion planners. We generate 100 random path planning scenarios. The objects are randomly placed on the surface of the table, then use the kinematics feasibility policy for problem ${\cal P}_1$ and only evaluate different choices for motion planners for problem ${\cal P}_2$. The results in Fig. \ref{fig:two_robot} (Right) show that BITstar and ABITstar perform the best. In summary, this can be explained because these two motion planners' internal heuristics tend to first plan from start to goal in a straight line, assuming, just like in our case, no obstacles. When the first solution is found, it will be refined until convergence. As a result, IMAPIP uses BITstar as the main motion planner.

\section{Experiments and Results}
We now describe our experiment setting and main results.

\subsection{Experimental Setup}
Our setup are two and four simulated Franka-Emika Panda robots where each robot has 7-degrees of freedom (DOF). We use PyBullet \cite{coumans2021} to build IMAPIP. The two-arm picking setting places two robots sit side to side, facing front and at a one-meter distance, while the four-arm picking places extra two robots on the other side of the table to the two-arm setting. An RGB-D camera with a top view of the bin that is located in the center of the table.

The objects used during the simulation comes from the YCB object dataset \cite{DBLP:journals/corr/CalliWSSAD15}. The 10 objects used in the simulation were a pair of the list of each of following objects: a banana, a barge clamp, a screwdriver, a wrench, and a knife. The objects lie on a flat surface with a high level of clutter, meaning that objects can overlap, be in contact, or be on the top of each other. The objects are placed in one of these two configuration: centred object placement for object that distributed around the mid-axis between the two robot (or the mid-point of the four robots) and excentric object placement where the objects are distributed with a distance of 10 cm from the mid-axis. 



We evaluate the following four experiment settings:
\begin{itemize}
\item \emph{Picking with centred object placement}: This will evaluate the number of total trials to clear bins with centred object placement, for the two-robot setting. The split space and the kinematics feasibility policy, and a purely sequential one-robot picking approach are compared on 10 centred placed bins containing 10 objects each over 10 trials.
\item \emph{Picking with excentred object placement}: This tests the robustness of the split space and the geometry-aware policy using 10 excentred bins containing 10 objects each over 10 trials.
\item \emph{Evaluation of best-policy variations}: This compares the performance of the geometry-aware policy variations; for centred object placement for a total of 10 centred bins containing 10 objects each over 10 trials. 
\item \emph{Evaluation of four-arm picking}: This evaluates IMAPIP for a four-robot setting using the kinematic feasibility policy to demonstrate the scalability. As seen in Fig. \ref{fig:four_robot}, we make the robots grasp four objects randomly placed on the table where each robot can reach at least three objects for a total of ten different object placements. The robots are placed at the edges of an imaginary square of a size of one meter. We evaluate on 10 bins containing 4 objects each over 10 trials.
\end{itemize}

\subsection{Experiment Results}
\begin{figure*}[htp]
 \centering
 \includegraphics[width=0.24\textwidth]{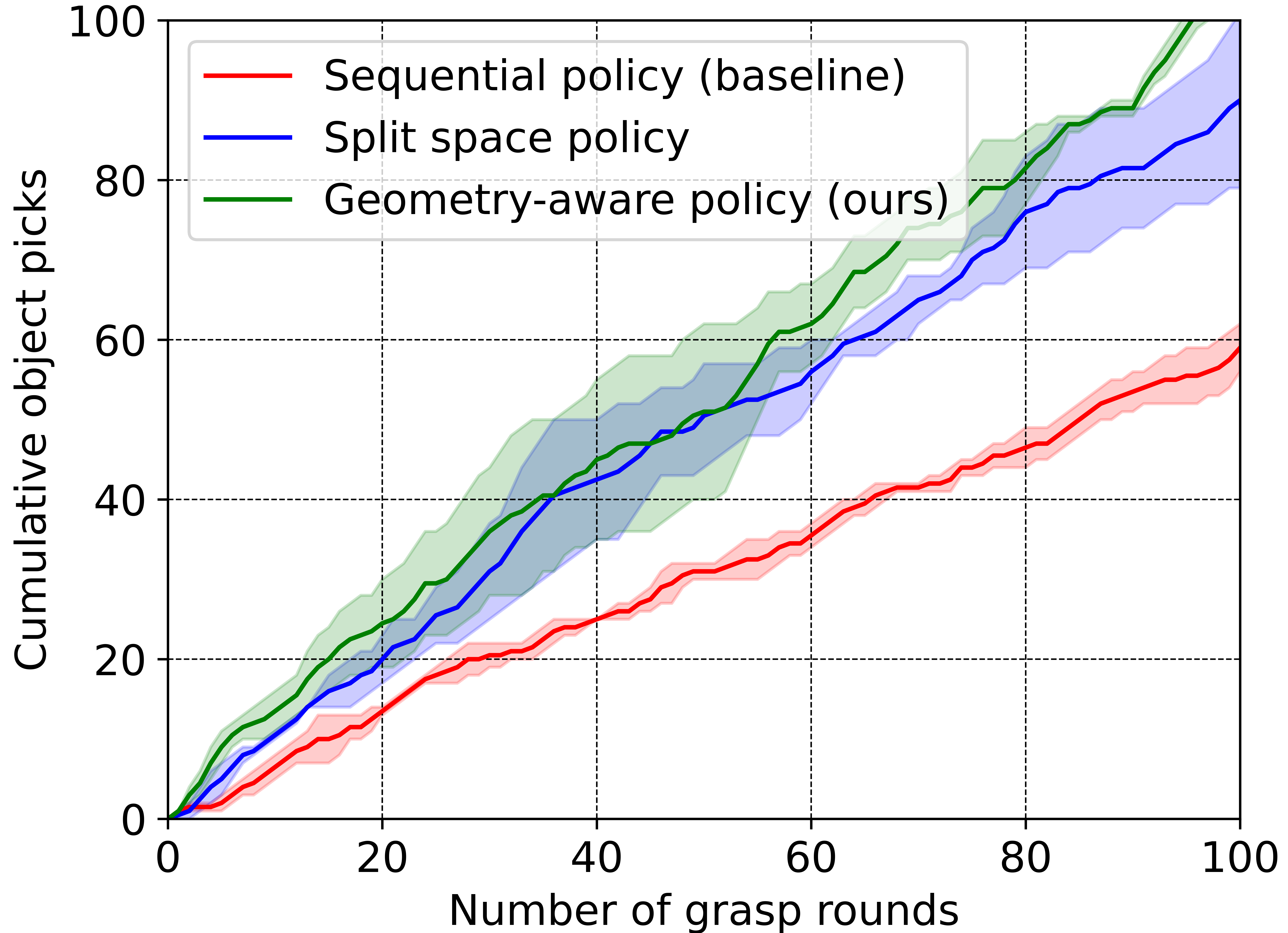}
 \includegraphics[width=0.24\textwidth]{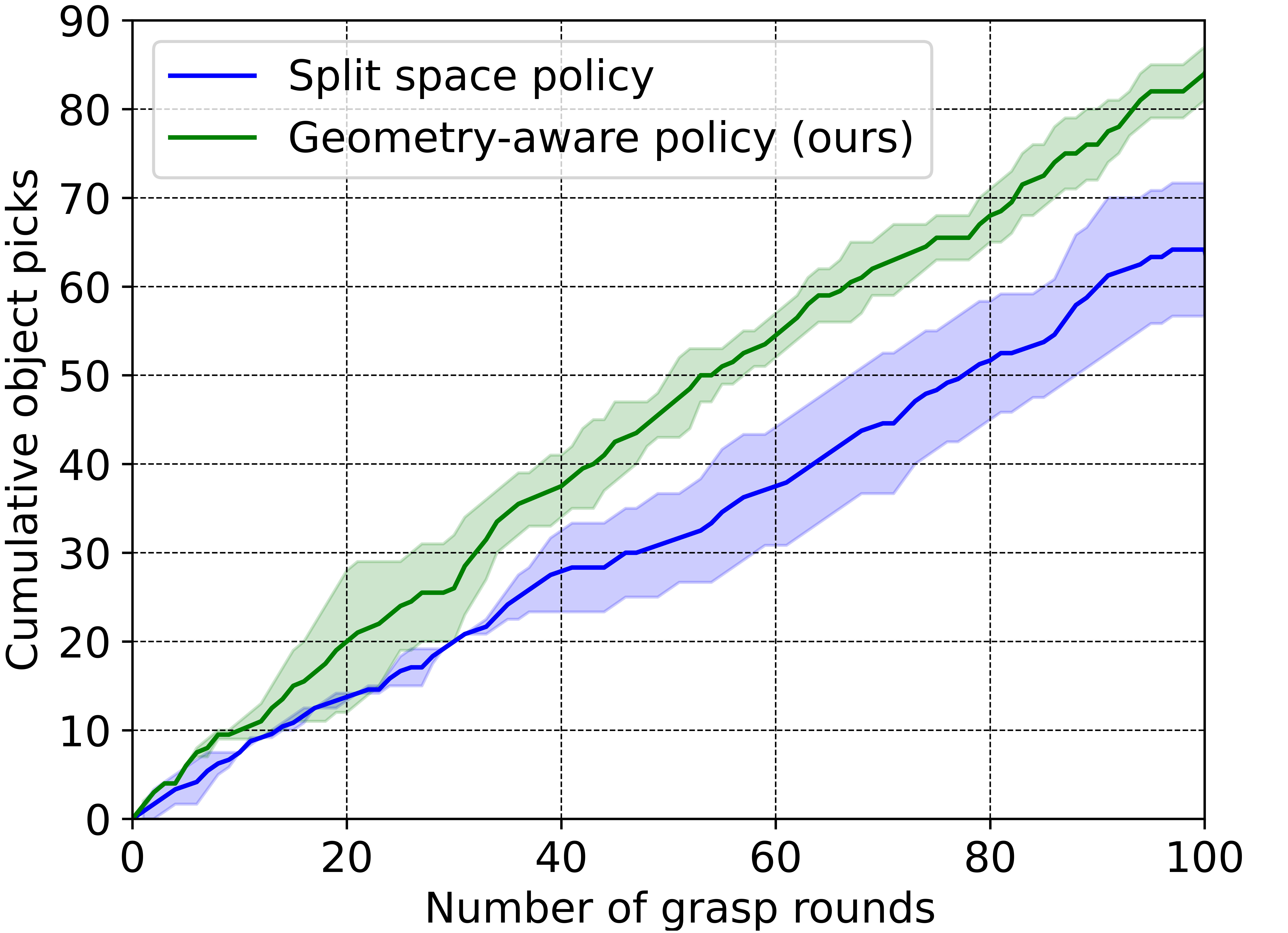}
 \includegraphics[width=0.24\textwidth]{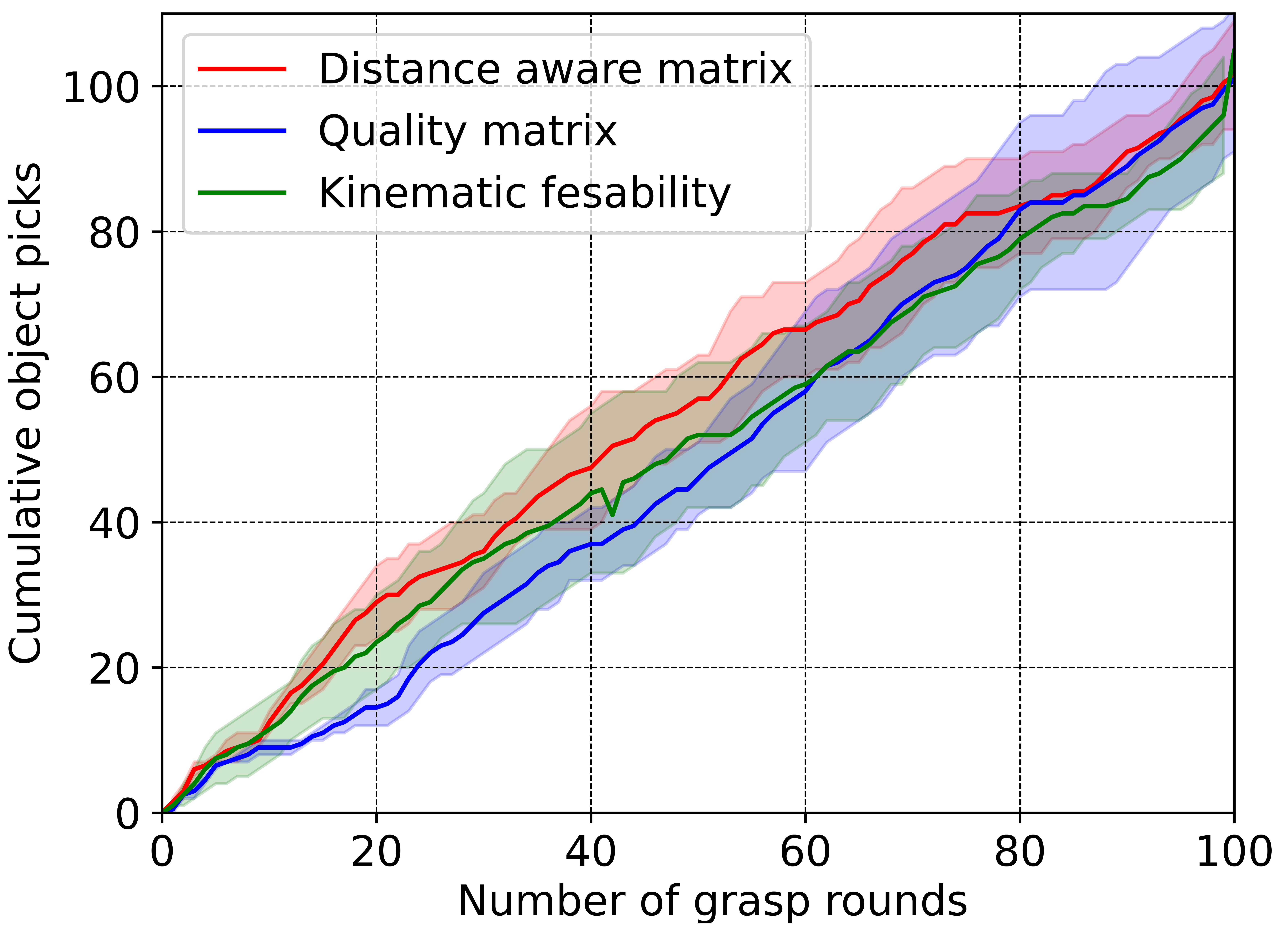}
 \includegraphics[width=0.24\textwidth]{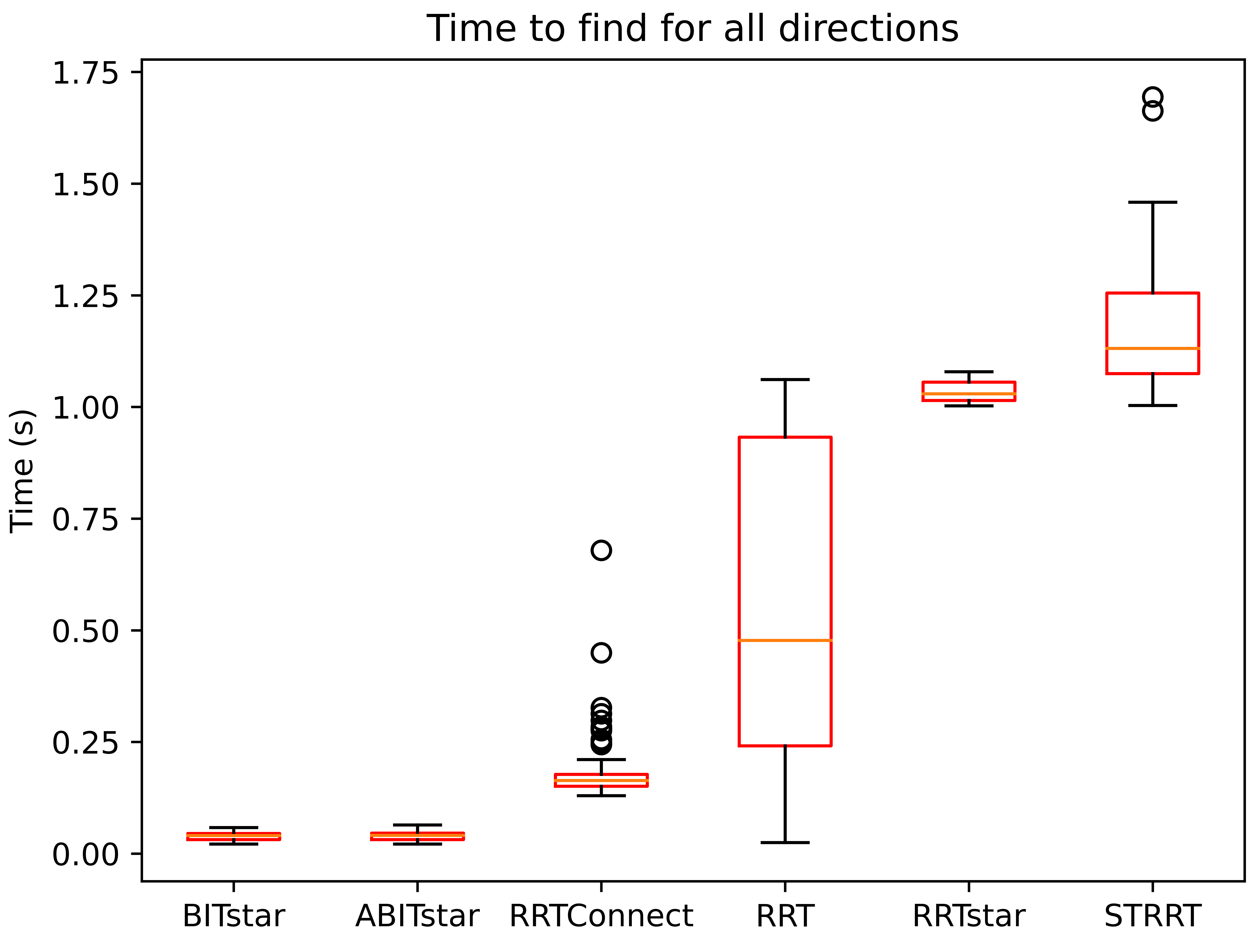}
 \caption{{\bf Left}): Evaluations of picking for centred object placement; {\bf Center-left}) Evaluations of picking for excentred object placement; {\bf Center-right}) Ablation study for the variation of the geometry-aware policy policies {\bf Right}) Benchmarking results for motion planning time, using the Open Motion Planning Library \cite{sucan2012the-open-motion-planning-library}.}
 \label{fig:two_robot}
\end{figure*}

\paragraph{Picking with centred object placement}
Figure \ref{fig:two_robot} (Left) reports the mean number of successful object picks over each picking round, where the shading areas represents the confidence intervals of 95\%.
The result shows that a dual-arm system using either the split or a kinematics feasibility policy increases the picking rate by approximately 28\% in comparison to a sequential policy. In overall, our proposed kinematics feasibility policy performs slightly better than the baseline split space policy, where it clears 10 bins of 100 objects with a fewer total number of picks (Two picks are counted in one grasp round if both robots simultaneously execute the assigned grasps).




\paragraph{Picking with excentred object placement} 
Figure \ref{fig:two_robot} (Center-left) depicts the averaged successful picks with a confidence intervals of 95\%. In this case, the bin is slightly shifted 10 centimeters to the right. We can observe that the kinematics feasibility policy performs significantly better than the split space space policy with approximately 31\% improvement. 

\paragraph{Evaluation of geometry-aware policy variation} 
Overall, to pick 100 objects of total 10 bins, the kinematic feasibility, distance aware, and grasp quality aware policy needed approximately 86, 88, and 88 grasp rounds, respectively. 
Even if this experiment does not show measurable results, the geometry-aware policy variation presents interesting characteristics as it can be adapted for different scenarios. For example, if our system wants to minimize the risk of robot collision or minimize the planning time, the distance-aware policy variation would fit the best. If we would like to pick the object with the grasp quality-aware policy variation would fit the best. 

\begin{figure*}[htp]
 \centering
 \includegraphics[width=0.23\textwidth]{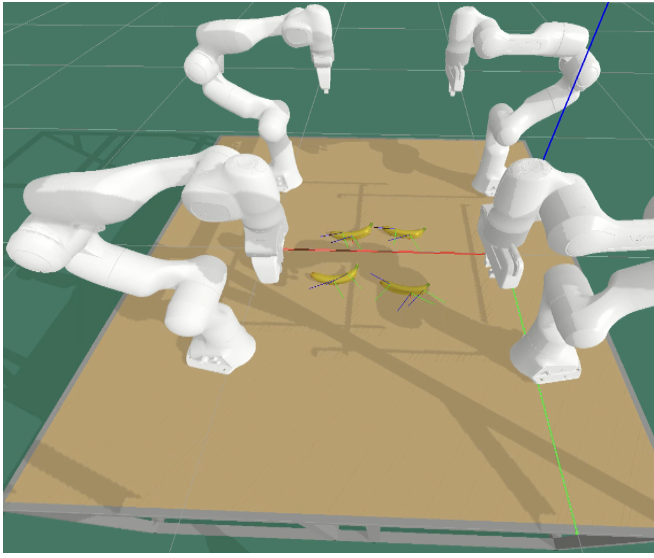}
 \includegraphics[width=0.23\textwidth]{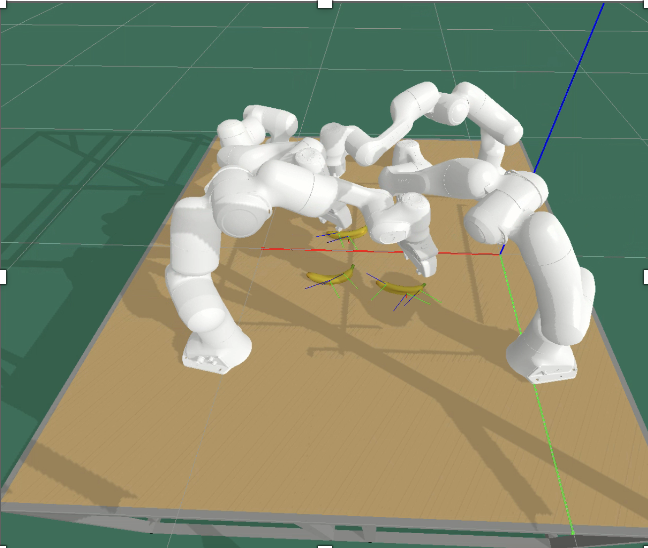}
 \includegraphics[width=0.23\textwidth]{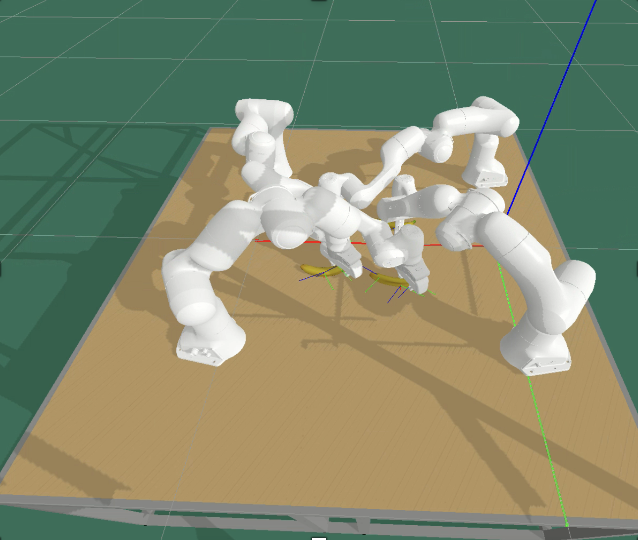}
 \includegraphics[width=0.23\textwidth]{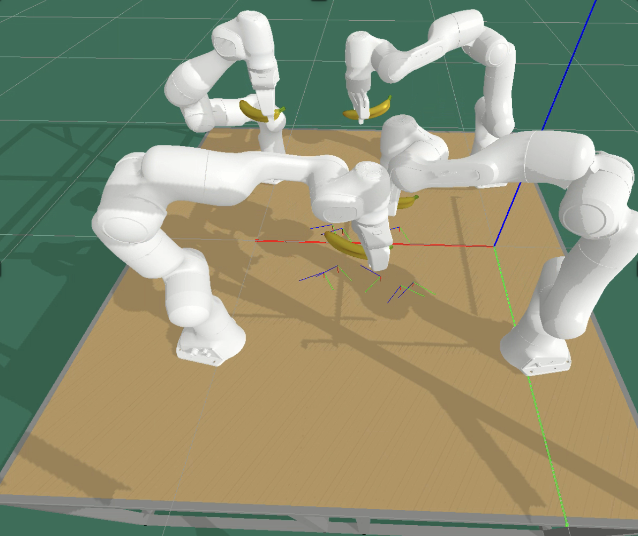}
 \caption{One typical execution sequence of four robot picking on a shared bin}
 \label{fig:four_robot}
\end{figure*}

\paragraph{Evaluation of four-arm picking}
We demonstrate that IMAPIP can scale for four robots. The success rate is 89.5\% using only one grasp round (where all four robots are used simultaneously). The cycle time to perform task and motion planning takes 606 ms where 90\% of this time is used to compute a solution for task planning using our geometry-aware policies as shown in Table \ref{tab:percentage_used_robot}. 
Figure \ref{fig:four_robot} shows a typical example where we can see that robots execute complex and close motion while picking the object without colliding. This result indicates that our system could scale well to multiple robots working on the same bin.

\begin{table}[]
\centering
\begin{tabular}{c|ccc|}
\cline{2-4}
                                & \multicolumn{3}{c|}{Four robot scenario}                                                                                                                                                                                              \\ \hline
\multicolumn{1}{|c|}{Time (ms)} & \multicolumn{1}{c|}{\begin{tabular}[c]{@{}c@{}}Motion \\ planning\end{tabular}} & \multicolumn{1}{c|}{\begin{tabular}[c]{@{}c@{}}Task \\ planning\end{tabular}} & \begin{tabular}[c]{@{}c@{}}Task and motion \\ planning\end{tabular} \\ \hline
\multicolumn{1}{|c|}{Mean}      & \multicolumn{1}{c|}{59.1}                                                       & \multicolumn{1}{c|}{546.9}                                                    & 605.0                                                               \\ \hline
\end{tabular}
\caption{Computation time to solve the POMDP-TAMP problem for four-robot setup.}
\label{tab:percentage_used_robot}
\end{table}


\subsection{Robot Usage Comparison}
We provide additional results to show the benefit of using our proposed IMAPIP using geometry-aware policy. Table \ref{tab:percentage_used_robot} shows that when object placement is centric, the baseline split space policy and the kinematics feasibility policies did not find a solution using two robots with 34.4\% and 37\% of the time, respectively. However, for the excentric scenario case the split space policy did not find a solution almost 60\% of the time while the kinematics feasibility policy was not affected by the change in the object placement and found the solution 61.4\% of the time. This means that the kinematics feasibility policy is more robust regarding object placement relative to the robot bases, which is a common scenario in industrial setups.

\begin{table}[]
\centering
\begin{tabular}{@{}c|ccc@{}}
\begin{tabular}[c]{@{}c@{}}Object \\ position\end{tabular} & \begin{tabular}[c]{@{}c@{}}Number of \\ robot used\end{tabular} & \begin{tabular}[c]{@{}c@{}}Split space \\ policy\end{tabular} & \begin{tabular}[c]{@{}c@{}}Kinematics feasibility \\ policy\end{tabular} \\ \midrule
\multirow{2}{*}{Centric}                                   & One robot                                                       & 34.4                                                          & 37.0                                                         \\
                                                           & Two robots                                                      & 65.6                                                          & 63.0                                                         \\ \midrule
\multirow{2}{*}{Excentric}                                 & One robot                                                       & 59.4                                                          & 38.6                                                         \\
                                                           & Two robots                                                      & 40.6                                                          & 61.4                                                        
\end{tabular}
\caption{Robot usage comparison for all pick attempts. Note that when a policy uses one robot means that there were no grasp feasible with two robots.}
\label{tab:percentage_used_robot}
\end{table}


\section{Conclusions}

In this work, we presented IMAPIP, an Integrated Multi-Arm bin-PIcking Pipeline using a task and motion planning under partial observability formulation. This pipeline is used as a benchmark platform to test and compare time to clear a bin for robotic grasping to maximize the number of picks per unit of time. We aim to reduce the computation of solving for a joint policy of robot-grasp assignment and motion planning in order to enable applications for real-time bin-picking. We have both applied several approximation techniques from previous TAMP work, and proposed reductions to task planning. The reduction results in different task planning policies, called geometry-aware policies. The geometry-aware policies solve for a skeleton of joint actions that are then used as fixed constraints for the motion planner.


Future work could address the limitation of our work that are summarized as follows. First, transferring to real robots is the most important possible extension of this work. Second, in the geometry-aware policies, additional heuristics could be added and combined to take advantage of their strength. For example, distance-aware and grasp quality-aware policy could be combined using a trade-off coefficient. Third, we could extend this work to other applications such as reaching from sorting objects for recycling to logistic presorting.

{
\balance 
\small 
\bibliography{references}
}

\end{document}